\documentclass[runningheads]{llncs}
\usepackage[T1]{fontenc}
\usepackage{graphicx,verbatim}
\usepackage{amsmath}
\usepackage{amssymb}
\usepackage{subfigure}
\usepackage{booktabs} 
\usepackage{multirow} 
\usepackage{tabulary}
\usepackage{url}
\usepackage{marvosym}
\begin{document}
%

\title{Co-Seg: Mutual Prompt-Guided Collaborative Learning for Tissue and Nuclei Segmentation}

\titlerunning{Co-Seg: Mutual Prompt-Guided Collaborative Learning for Segmentation}
%

\author{Qing Xu\inst{1,2}, 
Wenting Duan\inst{1}, \and 
Zhen Chen\inst{3}\textsuperscript{(\Letter)} 
}

%
\authorrunning{Q. Xu et al.}
%
\institute{University of Lincoln, Brayford Pool, UK \and
University of Nottingham, Nottinghamshire, UK \and
Yale University, New Haven CT 06510, USA\\
\,\,\,\email{zchen.francis@gmail.com}}

%
    
\maketitle              
\begin{abstract} 
Histopathology image analysis is critical yet challenged by the demand of segmenting tissue regions and nuclei instances for tumor microenvironment and cellular morphology analysis. Existing studies focused on tissue semantic segmentation or nuclei instance segmentation separately, but ignored the inherent relationship between these two tasks, resulting in insufficient histopathology understanding. To address this issue, we propose a Co-Seg framework for collaborative tissue and nuclei segmentation. Specifically, we introduce a novel co-segmentation paradigm, allowing tissue and nuclei segmentation tasks to mutually enhance each other. To this end, we first devise a region-aware prompt encoder (RP-Encoder) to provide high-quality semantic and instance region prompts as prior constraints. Moreover, we design a mutual prompt mask decoder (MP-Decoder) that leverages cross-guidance to strengthen the contextual consistency of both tasks, collaboratively computing semantic and instance segmentation masks. Extensive experiments on the PUMA dataset demonstrate that the proposed Co-Seg surpasses state-of-the-arts in the semantic, instance and panoptic segmentation of tumor tissues and nuclei instances. The source code is available at \url{https://github.com/xq141839/Co-Seg}.


\end{abstract}
\section{Introduction}
Medical image segmentation plays a crucial role in clinical applications and has received extensive attention in the research \cite{guo2020complementary,shui2024unleashing,horst2024cellvit,cheng2024unleashing}. Particularly, histopathology image analysis is challenged by the delineation of tissue regions and further separation of individual nuclei within the regions to assess tissue subtypes and tumor grading \cite{zhao2024foundation}. These requirements have led to two significant segmentation tasks in the field, \textit{i.e.}, tissue semantic segmentation and nuclei instance segmentation. 

For medical semantic segmentation, existing methods \cite{ronneberger2015u,xu2023dcsau,rahman2024emcad,xu2025hrmedseg} rely on hierarchical encoding and decoding to enhance multi-scale awareness for precise mask generation. In particular, ViT-based \cite{chen2024transunet} and Mamba-based \cite{liu2024swin} architectures model long-range dependencies to gain global contexts of target regions. For instance segmentation, a series of works \cite{graham2019hover,stringer2021cellpose,chen2023cpp,horst2024cellvit} adopted different distance proxy maps to improve the boundary understanding of instances. Recent SAM-based methods \cite{chen2024sam,na2024segment,shui2024unleashing,cheng2024unleashing,shui2024unleashing,chen2024asi,xu2024esp,NEURIPS2024_50ee6db5,lou2025nusegdg} have revealed impressive medical semantic and instance segmentation performance by manually providing corresponding task prompts.

Despite the advancements, existing methods \cite{chen2024transunet,valanarasu2022unext,liu2024rolling,xing2024segmamba,hao2024emf} usually focus on the optimization of tissue semantic or nuclei instance segmentation tasks, which can only be supervised by the isolated knowledge. In fact, both tasks are highly correlated as they aim to achieve the adequate perception and understanding of histopathology images. For example, accurately identifying nuclei provides valuable cues for understanding the underlying tissue structures, while tissue segmentation can aid in localizing nuclei. Therefore, this strong interdependence motivates us to develop a collaborative approach that integrates tissue and nuclei segmentation to advance state-of-the-art histopathology image analysis.

\begin{figure}[!t]
  \centering
  \includegraphics[width=1\linewidth]{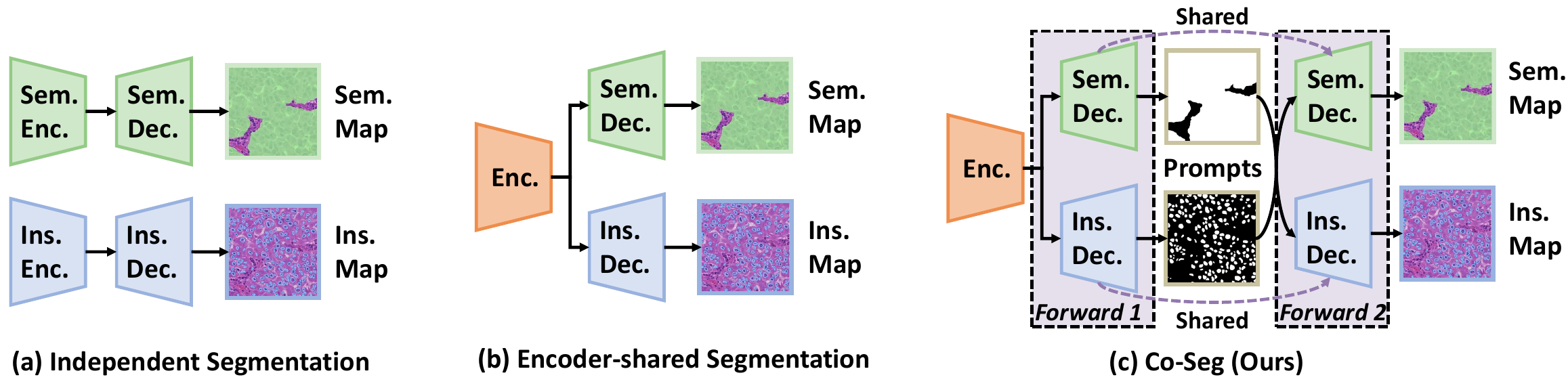}
  \caption{Comparison of our Co-Seg and existing works in tissue and nuclei segmentation. (a) Two independent networks for tissue and nuclei segmentation. (b) A shared image encoder but separated task decoders for tissue and nuclei segmentation. (c) Our Co-Seg leverages mutual prompts for collaborative tissue and nuclei segmentation.}
  \label{fig:intro}
\end{figure}

To overcome this bottleneck, we propose \textbf{Co-Seg}, a collaborative tissue and nuclei segmentation framework that allows semantic and instance segmentation tasks to mutually enhance each other. As illustrated in Fig. \ref{fig:intro}(c), the Co-Seg, based on a co-segmentation paradigm, improves segmentation mask quality by capturing contextual dependencies between the two tasks. Specifically, we first devise the region-aware prompt encoder (RP-Encoder) to extract high-quality semantic and instance prompts from the target region of both tasks, guiding segmentation decoding. We further introduce a mutual prompt mask decoder (MP-Decoder), which jointly computes semantic and instance maps by leveraging cross-guidance. This collaborative learning approach achieves contextual consistency, reducing prediction errors in both tasks. Experimental results on melanoma tissue and nuclei segmentation tasks demonstrate that our Co-Seg achieves remarkable performance over state-of-the-art segmentation methods.

\begin{figure}[!t]
  \centering
  \includegraphics[width=1\linewidth]{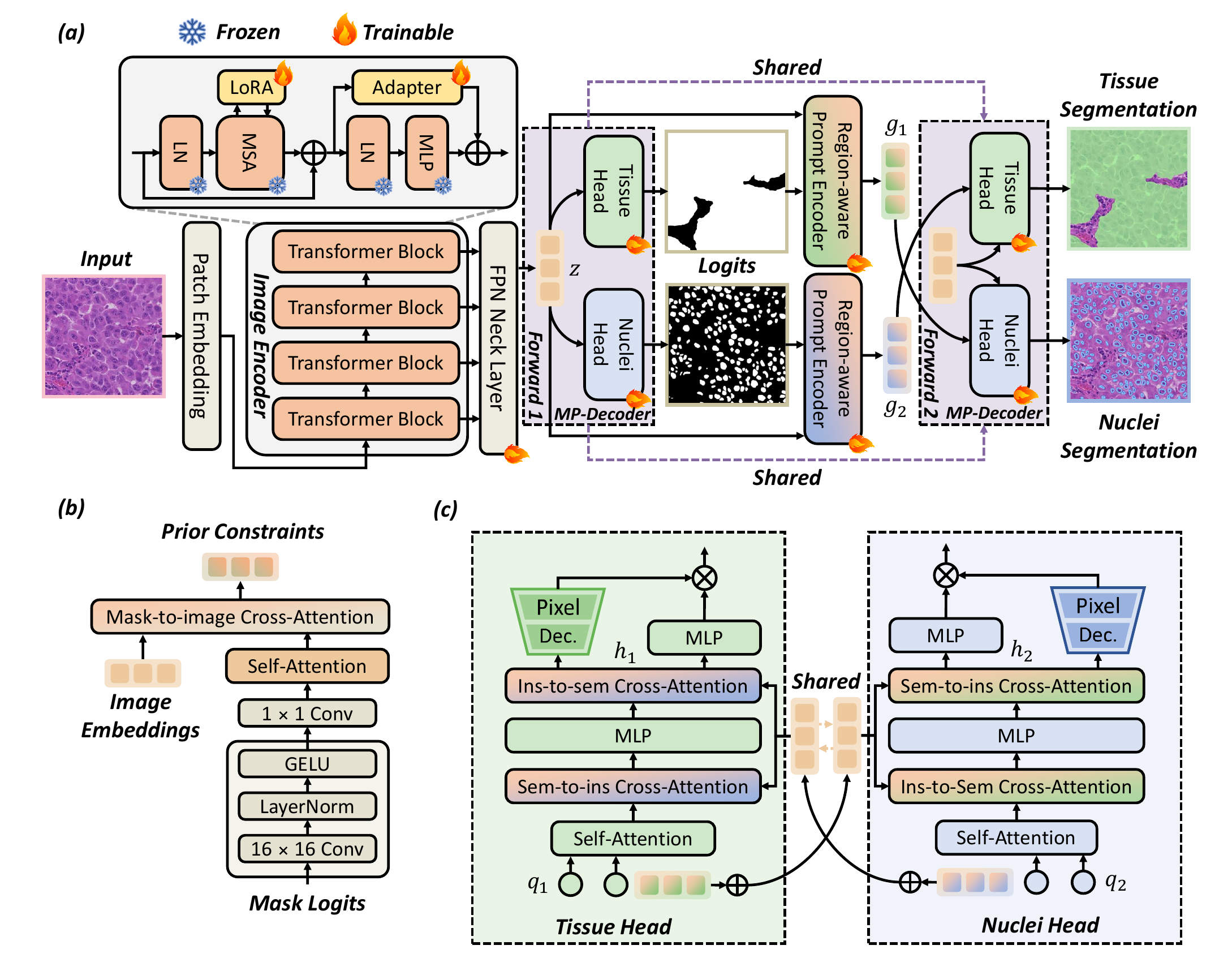}
  \caption{(a) The overview of the proposed Co-Seg framework for collaborative tissue and nuclei segmentation, consisting of (b) RP-Encoder and (c) MP-Decoder. Co-Seg fully exploits complementary information by leveraging mutual prompts.}
  \label{fig:method}
\end{figure}

\section{Methodology}
As elaborated in Fig. \ref{fig:method}, we devise the Co-Seg framework with the co-segmentation paradigm, achieving the mutual optimization between tissue and nuclei segmentation in medical imaging. To accomplish this, we design the RP-Encoder to provide prior constraints to guide both tasks, and the MP-Decoder to cooperatively generate semantic and instance maps through bidirectional information interactions. This integrated approach enables the two tasks to enhance each other, leveraging mutual prompts for improved segmentation accuracy.

\subsection{Co-Segmentation Paradigm}
Existing tissue \cite{he2023h2former,ibtehaz2023acc,nam2024modality} and nuclei \cite{nam2023pronet,meng2024nusea,chen2024sam} segmentation methods decouple the parameter space of semantic and instance segmentation, disrupting their interdependencies, and degrading the performance. To address this issue, we propose the co-segmentation paradigm that leverages closed-loop bidirectional interaction to realize dual-task collaborative optimization. Specifically, our goal is to train a model $f_\Theta: x \rightarrow \{y_{1}, y_{2}\}$, where $\Theta=\{\theta_{1}, \theta_{2}\}$ represents the parameters for semantic segmentation $\theta_1$ and instance segmentation $\theta_2$, with $y_{1}$ and $y_{2}$ as the corresponding segmentation masks. To model the interdependence between the two tasks, we define their joint prediction of $y_1$ and $y_2$ as:
\begin{equation} \label{eq:1}
p(y_1, y_2 | x, \theta_1, \theta_2) = p(y_1 | x, \theta_1) \cdot p(y_2 | y_1, x, \theta_2) = p(y_2 | x, \theta_2) \cdot p(y_1 | y_2, x, \theta_1).
\end{equation}
Note that Eq. \eqref{eq:1} reveals the mutual dependency between semantic and instance segmentation through intertwined conditional probabilities. It states that $p(y_1, y_2 | x, \theta_1, \theta_2)$, the joint probability of obtaining both segmentations given the image and model parameters, can be decomposed in two symmetrical ways:
\begin{itemize}
  \item $p(y_1 | x, \theta_1) \cdot p(y_2 | y_1, x, \theta_2)$ suggests that once the semantic segmentation $y_1$ is known, it directly influences the prediction of the instance segmentation $y_2$.
  \item Similarly, $p(y_2 | x, \theta_2) \cdot p(y_1 | y_2, x, \theta_1)$ implies that knowing the instance segmentation $y_2$ affects the outcome of the semantic segmentation $y_1$.
\end{itemize}
Both expressions are mathematically equivalent because they describe the same joint probability from different perspectives, highlighting that each segmentation task provides crucial context that enhances the accuracy of the other. This reciprocal relationship underscores the need for a collaborative approach, where both segmentation tasks are optimized together rather than in isolation, leveraging the full context available from each task to improve overall segmentation results. Translating this relationship to the gradient calculations for $\theta_1$ and $\theta_2$, we have:
\begin{equation} \label{eq:2}
\nabla_{\theta_i} \ell_i = -\nabla_{\theta_i} \mathbb{E} \left[ \log p(y_i | x, \theta_i) \right] - \nabla_{\theta_i} \mathbb{E} \left[ \log p(y_j | y_i, x, \theta_j) \right] + \nabla_{\theta_i} I(y_1; y_2 | x),
\end{equation}
for $i, j \in \{1,2\}, i \neq j$ where $\nabla_{\theta} I(y_1; y_2 | x) \approx \mathbb{E} \left[ \nabla_{\theta} \log \frac{p(y_1, y_2 | x)}{p(y_1 | x)p(y_2 | x)} \right]$. This term captures the mutual information between 
$y_1$ and $y_2$, ensuring that both segmentation tasks benefit from shared feature learning. The first term of Eq. \eqref{eq:2} represents the main segmentation loss gradient, while the second term incorporates the implicit gradient of task interdependencies, modeling the complementarity between $\theta_1$ and $\theta_2$. This leads to the following optimization update rule:
\begin{equation}
    \theta_1 \leftarrow \theta_1 - \eta \nabla_{\theta_1} \ell_1, \quad \theta_2 \leftarrow \theta_2 - \eta \nabla_{\theta_2} \ell_2,
\end{equation}
where $\eta$ is the learning rate. In this way, the co-segmentation paradigm breaks the barrier of isolating gradient flows from each other in multi-task learning, enabling our Co-Seg framework to achieve the bidirectional optimization of tissue and nuclei segmentation.

\subsection{Region-aware Prompt Encoder}
The proposed co-segmentation paradigm requires establishing implicit interdependencies between tissue and nuclei segmentation tasks. To achieve this, we introduce the RP-Encoder that leverages mask prompts to capture task-specific target regions as prior constraints, as presented in Fig. \ref{fig:method}(b). Specifically, we first employ several convolutional blocks to extract features $m_i$ from mask logits $\mu_i$. After that, the RP-Encoder performs self-attention, followed by cross-attention with shared image embeddings $z$ to generate a set of dense prompts $g_i$, as follows:
\begin{equation}
m_i = {\rm Conv_2(GELU(LN(Conv_1}(\mu_i)))),
\end{equation}
\begin{equation}
g_i = {\rm CrossAttention(SelfAttention}(m_i), z, z),
\end{equation}
where $\rm Conv_1$ denotes a $16 \times 16$ convolutions for downsampling, $\rm Conv_2$ is a $1 \times 1$ convolution for channel expansion, $\rm LN(\cdot)$ is LayerNorm. These operations encode the structural context of mask logits. On this basis, we apply RP-Encoder to semantic and instance mask logits, effectively obtaining region-aware prompts: $g_1$ and $g_2$ as prior constraints of Co-Seg for tissue and nuclei segmentation tasks.  

\subsection{Mutual Prompt Mask Decoder}
Following the co-segmentation paradigm, we propose the MP-Decoder to enforce bidirectional interactions between tissue and nuclei segmentation, ensuring mutual refinement rather than treating them as isolated tasks, as illustrated in Fig. \ref{fig:method}(c). Specifically, the MP-Decoder contains tissue and nuclei heads that leverage two sets of query embeddings: $q_1$ and $q_2$ to save the decoding information of both tasks. To leverage complementary effects between tasks, we first perform self-attention on each query, followed by cross-attention with the other task’s prompts, as follows:
\begin{equation}
q'_i = {\rm CrossAttention(SelfAttention}(q_i), z\oplus g_j, z\oplus g_j),
\end{equation}
where $\oplus$ represents the element-wise addition. This is followed by applying MLP to ensure the refined query embeddings are expressive. The MP-Decoder further applies the reverse cross-attention to generate task-specific image embeddings $h_1$ and $h_2$, as follows: 
\begin{equation}
h_i = {\rm CrossAttention}(z\oplus g_j, q'_i, q'_i).
\end{equation}
On this basis, nuclei context prompts can enhance tissue segmentation consistency, while tissue boundary feedback provides prior constraints for nuclei discrimination. Following the standard SAM \cite{kirillov2023_sam}, each segmentation head adopts a pixel decoder to upsample the refined task-specific image embedding. Finally, our MP-Decoder generates tissue and nuclei segmentation predictions by performing the dot product between their upscaled image embeddings and corresponding task queries. Overall, the proposed MP-Decoder utilizes the bidirectional interactions of tissue and nuclei segmentation information to mutually improve the quality of segmentation masks of our Co-Seg framework.

\subsection{Optimization Pipeline}
To construct our Co-Seg framework, we first adopt Hirea ViT \cite{ryali2023hiera} as the shared image encoder for tissue and nuclei segmentation, ensuring the consistency of feature learning. In particular, we load the pre-trained SAM2-L \cite{ravi2025sam} to initialize corresponding parameters and freeze these weights to preserve pre-trained knowledge. Additionally, we insert LoRA \cite{hu2022lora} and Adapter \cite{houlsby2019parameter} into attention and FFN layers to achieve parameter-efficient fine-tuning from natural to histopathology domains. Based on the co-segmentation paradigm, Co-Seg includes two decoding forwards, where the tissue and nuclei heads of the MP-Decoder share parameters in these two stages. Specifically, the MP-Decoder leverages the image embedding generated from the image encoder to calculate binary segmentation masks $y'_1$ and $y'_2$ of both tasks without any prompts. Then, we deliver them to the RP-Encoder to provide tissue and nuclei prompts. In the second forward, they will be directly transferred back to the MP-Decoder as prior constraints. Finally, the MP-Decoder utilizes cross-guidance to co-generate tissue and nuclei segmentation masks. The training of Co-Seg consists of two parts: (1) the optimization of tissue $g_1$ and nuclei $g_2$ prior constraints and (2) the joint optimization of tissue and nuclei segmentation. The overall loss is formularized as:
\begin{equation}
    \mathcal{L}_{\rm CoSeg} = \underbrace{\lambda_1\mathcal{L}_{\rm sem}^{g_1}(\hat{y}'_1,y'_1) + \lambda_2\mathcal{L}_{\rm ins}^{g_2}(\hat{y}'_2,y'_2)}_{\text{Prior Constraint Loss}} + \underbrace{\mathcal{L}_{\rm sem}(\hat{y}_1,y_1) + \mathcal{L}_{\rm ins}(\hat{y}_2,y_2)}_{\text{Segmentation Loss}},
\end{equation}
where $\lambda_1$ and $\lambda_2$ are factors to adjust the contribution of each term. By optimizing $\mathcal{L}_{\rm CoSeg}$, our Co-Seg achieves accurate tissue and nuclei segmentation with superior performance across different histopathology images.

\begin{table}[!t]
\begin{minipage}[!t]{0.49\textwidth}
\makeatletter\def\@captype{table}
\caption{Comparison with state-of-the-arts on tissue semantic segmentation.}
\scalebox{0.94}{\begin{tabular}{l|lll}
\hline
Methods & Dice & mIoU & HD \\
\hline
nnUnet \cite{isensee2021nnu} & 90.90  & 84.48 & 236.03 \\
Swin-Umamba \cite{liu2024swin} & 89.67  & 82.75  & 265.43 \\
EMCAD \cite{rahman2024emcad} & 90.96 & 84.61  & 306.36 \\
SAC \cite{na2024segment} & 91.19  & 85.01 & 295.57 \\
H-SAM \cite{cheng2024unleashing} & 91.47  & 85.47  & 246.16 \\
\hline
Co-Seg  & \textbf{92.51} & \textbf{87.18} & \textbf{206.72} \\
\hline
\end{tabular}}
\label{tab:tis}
\end{minipage}
\begin{minipage}[!t]{0.49\textwidth}
\makeatletter\def\@captype{table}
\caption{Comparison with state-of-the-arts on nuclei instance segmentation}
\scalebox{0.94}{\begin{tabular}{l|llll}
\hline
Methods & F1 & Prec. & Rec. & AJI \\
\hline
HoverNet \cite{graham2019hover} & 75.53 & 70.53 & 82.80 & 64.40 \\
CellPose \cite{stringer2021cellpose} & 75.30 & 71.12 & 81.79 & 65.15 \\
CPPNet \cite{chen2023cpp} & 74.96 & 69.58 & 83.06 & 64.63 \\
CellViT \cite{horst2024cellvit} & 76.04 & 70.21 & 83.51 & 66.13 \\
PromptNucSeg \cite{shui2024unleashing} & 76.47 & 71.81 & 83.02 & 66.71 \\
\hline
Co-Seg & \textbf{79.70} & \textbf{77.05} & \textbf{83.61} & \textbf{69.14} \\
\hline
\end{tabular}}
\label{tab:nuc}
\end{minipage}
\end{table}

\section{Experiments}
\subsection{Experimental Setup}
To validate the effectiveness of the proposed Co-Seg, we adopt the melanoma-specific histopathology dataset: PUMA \cite{giaf011} for tissue semantic and nuclei instance segmentation tasks. It contains 206 histopathology images of $1024 \times 1024$ resolution and adopts a common split of training, validation, and test sets as 7:1:2. We perform all experiments on a NVIDIA A5000 GPU
using PyTorch. For fair comparisons, we implement all tissue and nuclei segmentation methods with the same training settings and
configurations. We utilize the pre-trained SAM ViT-H \cite{kirillov2023_sam} structure as the image encoder of medical SAM baselines. We apply the optimizer using Adam with an initial learning rate of $1\times10^{-4}$ and use the exponential decay strategy to adjust the learning rate with a factor of 0.98. The batch size and epochs are set to 16 and 300. We leverage the combination of cross-entropy loss and Dice loss to supervise the tissue segmentation. For nuclei segmentation, we follow the standard combination loss of Focal loss, Dice loss, MSE loss and MSGE loss \cite{graham2019hover,horst2024cellvit}. The loss coefficient $\lambda_1$ and $\lambda_2$ are set to 2 and 1. Our Co-Seg introduces additional 9.1\% learnable parameters to the baseline (in Table \ref{tab:ablation}) during fine-tuning.


\begin{table}[!t]
\centering
\tabcolsep=0.2cm
\caption{Comparison with state-of-the-arts on histopathology panoptic segmentation.}
\label{tab:metrics-comparison}
\begin{tabular}{l|ccc|ccc}
\hline
Methods & PQ$_{\rm tissue}$ & DQ$_{\rm tissue}$ & SQ$_{\rm tissue}$ & PQ$_{\rm nuclei}$ & DQ$_{\rm nuclei}$ & SQ$_{\rm nuclei}$ \\
\hline
nnUnet \cite{isensee2021nnu} & 57.34 & 68.27 & 70.13 & 58.60 & 72.83 & 80.24 \\
HoverNet \cite{graham2019hover} & 58.95 & 67.25 & 69.92 & 60.91 & 74.96 & 81.04  \\
CellViT \cite{horst2024cellvit} & 60.55 & 71.32 & 69.17 & 62.91  & 76.59 & 81.95 \\
H-SAM \cite{cheng2024unleashing} & 61.62 & 71.17 & 70.81 & 62.68 & 76.39 & 81.86 \\
PromptNucSeg \cite{shui2024unleashing} & 61.58 & 70.84 & 70.92 & 63.37 & 77.41 & 81.79 \\
\hline
Co-Seg  & \textbf{63.09} & \textbf{72.82} & \textbf{70.99} & \textbf{66.11} & \textbf{79.70} & \textbf{82.80} \\
\hline
\end{tabular}
\label{tab:pan}
\end{table}

\begin{table}[!t]
    \centering
    \tabcolsep=0.26cm
    \caption{Ablation study of Co-Seg on PUMA tissue and nuclei Segmentation.}
    \label{tab:ablation}
    \begin{tabular}{lll|lll|llll}
        \hline
        \multirow{2}{*}{$P$} & \multirow{2}{*}{$D$} & \multirow{2}{*}{$C$} & \multicolumn{3}{c|}{Tissue Semantic Seg.} & \multicolumn{4}{c}{Nuclei Instance Seg.} \\
        \cline{4-10}
        & & & Dice & mIoU & HD & F1 & Prec. & Rec. & AJI \\
        \hline
        & & & 90.98 & 84.71 & 287.19 & 76.35 & 73.18 & 82.27 & 66.59 \\
        \checkmark & & & 91.23 & 85.18 & 261.37 & 77.01 & 73.25 & 83.01 & 67.08 \\
        \checkmark & \checkmark & & 91.76 & 86.13 & 246.56 & 78.19 & 74.91 & 82.54 & 67.93 \\
        \checkmark & \checkmark & \checkmark & \textbf{92.51} & \textbf{87.18} & \textbf{206.72} & \textbf{79.70} & \textbf{77.05} & \textbf{83.61} & \textbf{69.14} \\
        \hline
    \end{tabular}
\end{table}

\subsection{Comparison with State-of-the-art Methods}

\noindent \textbf{Tissue Semantic Segmentation.} We first evaluate the performance of all models in the tissue semantic segmentation. As illustrated in Table \ref{tab:tis}, we observe that classical semantic segmentation methods are inferior to medical SAMs, \textit{e.g.,} H-SAM \cite{cheng2024unleashing} surpasses EMCAD \cite{rahman2024emcad} with a 0.51\% Dice increase. Remarkably, our Co-Seg achieves the best performance with a P-value $< 0.005$, Dice of 92.51\%, and the lowest HD of 206.72, indicating precise tissue boundary localization.

\noindent \textbf{Nuclei Instance Segmentation.} The comparison of nuclei instance segmentation is shown in Table \ref{tab:nuc}. Our Co-Seg framework reveals the overwhelming performance of four metrics. In particular, Co-Seg achieves a significant advantage over the second-best PromptNucSeg \cite{shui2024unleashing}, \textit{e.g.,} a P-value $< 0.001$, a F1-score increase of 3.23\%, and an AJI increase of 2.43\%. 

\begin{figure}[!t]
  \centering
  \includegraphics[width=1\linewidth]{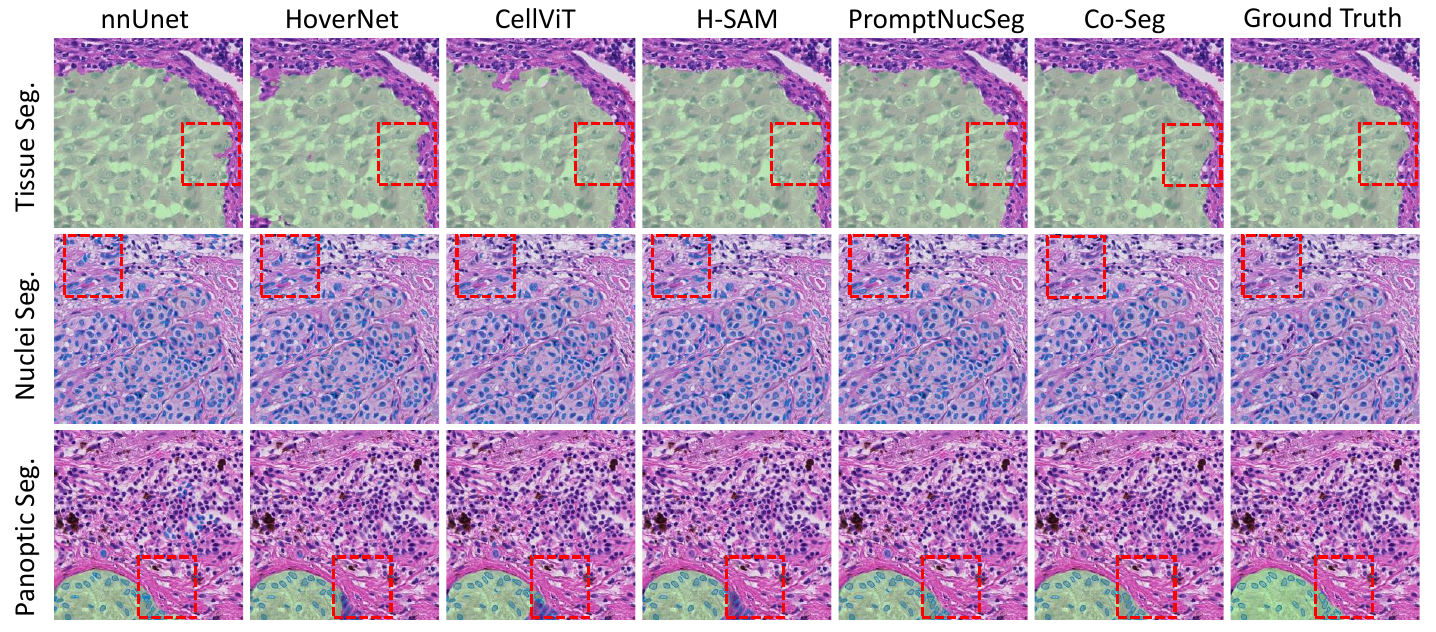}
  \caption{Qualitative comparison of three evaluation protocols on tissue semantic and nuclei instance segmentation. Benefiting from the mutual reinforcement of semantic and instance segmentation tasks, our Co-Seg can delineate precise tissue regions and segment accurate nuclei, containing fewer false positives.}
  \label{fig:visual}
\end{figure}

\noindent \textbf{Histopathology Panoptic Segmentation.} We further comprehensively evaluate the performance of our Co-Seg with the panoptic segmentation protocol. As presented in Table \ref{tab:pan}, our Co-Seg benefits from the novel co-segmentation paradigm, achieving the best performance of all six metrics, with a remarkable PQ increase of 1.51\% and 2.74\% in tissue and nuclei segmentation tasks. The qualitative comparison is visualized in Fig. \ref{fig:visual}.

\subsection{Ablation Study}
To investigate the effectiveness of our proposed co-segmentation paradigm $C$, RP-Encoder $P$ and MP-Decoder $D$, we further conduct the comprehensive ablation study on the tissue semantic and nuclei instance segmentation of the PUMA dataset, as illustrated in Table \ref{tab:ablation}. By removing the tailored modules from Co-Seg, we construct two independent Hiera ViT-based semantic and instance segmentation networks as the ablation baseline. By introducing the RP-Encoder, the performance achieves Dice and F1-score increases of 0.25\% and 0.66\% on tissue semantic and nuclei instance segmentation, respectively. Moreover, we investigate the effect of combined RP-Encoder and MP-Decoder, resulting in superior performance, with the Dice of 91.76\% and the F1-score of 78.19\%. Finally, we establish the framework using the proposed co-segmentation paradigm. The result proves that this design can significantly prompt the semantic and instance segmentation capabilities. In this way, these ablation experiments demonstrate the effectiveness of the RP-Encoder, MP-Decoder and co-segmentation paradigm in our Co-Seg framework.

\section{Conclusion}
In this work, we identify the potential relationship between tissue and nuclei segmentation in histopathology and propose a novel co-segmentation paradigm to establish the Co-Seg framework for promoting nuclei and tissue segmentation mutually.  It comprises two modules: the RP-Encoder aims to provide task-specific region prompts by perceiving the target area, and the MP-Decoder adopts cross-guidance to cooperatively generate tissue and nuclei segmentation masks. Extensive experiments on the melanoma tissue and nuclei instance segmentation dataset demonstrate that Co-Seg outperforms existing methods by remarkable margins.

\subsubsection{Disclosure of Interests.} The authors declare no competing interests.

\bibliographystyle{splncs04}
\bibliography{Paper-0931}

\end{document}